\documentclass[]{article}
\usepackage[utf8]{inputenc}

\usepackage{blindtext}
\usepackage{multicol}
\usepackage[normalem]{ulem}
\usepackage[paper=portrait,pagesize]{typearea}
\usepackage{soul}
\usepackage{geometry}
\usepackage{lipsum} 
\usepackage{pdfpages}
\usepackage{pdflscape}
\usepackage{subcaption}
\usepackage{hyperref}

\usepackage{adjustbox}
\usepackage{multirow, booktabs}
\usepackage{longtable}
\usepackage{makecell, cellspace, caption}

\hypersetup{
    colorlinks,
    citecolor=blue,
    filecolor=black,
    linkcolor=black,
    urlcolor=blue
}
\DeclareCaptionFormat{custom}
{
    \textbf{#3}
}
\definecolor{gr}{rgb}{0.4, .8, 0.5}

\UseRawInputEncoding

\newcolumntype{L}[1]{>{\raggedright\let\newline\\\arraybackslash\hspace{0pt}}m{#1}}
\newcolumntype{C}[1]{>{\centering\let\newline\\\arraybackslash\hspace{0pt}}m{#1}}
\newcolumntype{R}[1]{>{\raggedleft\let\newline\\\arraybackslash\hspace{0pt}}m{#1}}

\title{Bayesian Network Models of Causal Interventions\\ in Healthcare Decision Making: \\ Literature Review and Software Evaluation}

\author{Artem Velikzhanin, Benjie Wang and Marta Kwiatkowska\\ \\
Technical Report\\ \\
Department of Computer Science\\ University of Oxford\\ \\
\\ \\ \\ \\ 
{\includegraphics[width=32mm]{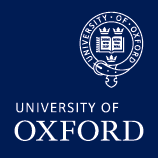}}
}

\begin{document}

\maketitle
%\begin{center}%{\includegraphics[width=32mm]{oxlogo}}
%\end{center}

%\includepdf[pages={-}]{pdf/title_1.pdf}
\newpage
\tableofcontents
\newpage

\begin{abstract}
This report\footnote{
This report was funded by the ERC under the European Union’s Horizon 2020 research and innovation programme (FUN2MODEL, grant agreement No.834115).} summarises the outcomes of a systematic literature search to identify Bayesian network models used to support decision making in healthcare. After describing the search methodology, the selected research papers are briefly reviewed, with the view to identify publicly available models and datasets that are well suited to analysis using the causal interventional analysis software tool developed in~\cite{Wang2021}. Finally, an experimental evaluation of applying the software on a selection of models is carried out and preliminary results are reported.
\end{abstract}

\section*{Introduction}
\addcontentsline{toc}{section}{Introduction}
In the medical field, prediction of the risk of disease requires establishing a statistical model of risk factors and
diseases, and prediction of the probability of disease according to levels of multiple risk factors.
Bayesian network models are often used in healthcare settings to support decision making, such as outcome prediction and disease management. Since Bayesian network models can support causal inference and interventions, algorithms and software tools have been developed, for example~\cite{Wang2021}, that can compute provable robustness guarantees and automate ``what if'' analysis of disease outcomes. However, in view of data often being proprietary and privacy concerns, there is a lack of publicly available models and datasets on which to evaluate the potential of these methods in healthcare settings.

This report summarises a systematic review of existing literature on Bayesian network models in medical settings undertaken as part of the FUN2MODEL project\footnote{\url{fun2model.org}}, with a specific focus on decision support that is based on causal interventions. A search methodology was developed to identify articles that meet a set of criteria, which were then reviewed to summarise the use cases, including the medical domain and effects of causal interventions, and insight obtained; those articles that provide publicly available models and datasets were highlighted. For a selection of models, causal Bayesian network representations were built and analysed using the IntRob software~\cite{Wang2021} \footnote{\url{https://github.com/wangben88/interventional-robustness}}, and preliminary results reported.

\section*{Literature Search Strategy and Analysis}
\addcontentsline{toc}{section}{Literature Search Strategy and Analysis}
This review is focused on identifying existing datasets and Bayesian network (BN) models that capture decision making in healthcare settings, and particularly those where the decision can be modelled as a causal intervention in the system.
Thus, articles were selected based on the following criteria:
\begin{itemize}
    \item[--] availability of the (Bayesian network) model;
    \item[--] availability of (categorical) data;
    \item[--] appropriate targets for intervention (for example, administering a drug);
    \item[--] clear intervention interpretation to determine the appropriate intervention bounds;
    \item[--] an appropriate target probability (e.g. survival of patient).
\end{itemize}

A comprehensive search of health and health informatics literature databases, including PubMed, ScienceDirect and Scopus, was performed using keywords arranged in the following general search query:\hfill \break

\textit{bayes AND (network OR model) AND (medical OR clinical) AND decision-making AND causal}\hfill \break

Terms such as `Bayesian networks' and `decision making' are included as they are widely used in the target literature. We also tried using `intervention', but this proved to be less useful as intervention is also a medical term.
In particular, in medicine, an intervention is a treatment, procedure, or other action taken to prevent or treat disease, or improve health in other ways \cite{national_cancer_institute}. As for interventions in BNs, in Woodward's theory \cite{Woodward2003} one variable X is a direct cause of another variable Y if there exists an intervention on X such that, if all other variables are held fixed at some value, X and Y are associated. Such an account assumes a lot about the sort of intervention needed, however, and Woodward goes to great lengths to make the idea clear \cite{Eberhardt2007}. 
A preliminary search showed that there is a small amount of work in this specific area, so the term `causal' was chosen as the most universal one, encompassing also the interventions that we are interested in.
Due to the large number of articles, we limited the selection to the areas of Computer Science, Psychology, Medicine and Dentistry, and Neuroscience, using filters. 

%As a result of the analysis, several articles were selected outside of 2012-2022. 
After extracting relevant articles based on the above criteria, we analyzed each article using natural language processing (NLP) tools to answer the following questions: in what medical areas are Bayesian network models used, and, in particular, what were the effects of causal interventions studied? For each article, we had access to the following attributes: title, abstract, authors, year, and DOI. The analysis was performed on the basis of available abstracts of articles. As of August 4, 2022, there were 1042 (983 in ScienceDirect; 24 in Scopus; 35 in PubMED) unique article abstracts available to download in ScienceDirect, Scopus and PubMED. 

The entire procedure/algorithm for the analysis of the articles is shown in Appendix A in Figure A.1.
In the first step, the PMIDs of the PubMED articles were converted to the BibTeX format. Then, DOI links from all three databases were converted into the same format and merged, allowing us to remove duplicate articles from the database.
For the remaining articles, we extracted the article abstracts; 29 abstracts were not available. 

Next, all raw abstracts were cleaned/preprocessed in order to faciliatate further analysis. We used regular expressions (combinations of wildcard characters’ operators for searching and manipulating substrings in text) to remove all parentheses in abstracts, and tokenize the words. Tokenization allows us to decompose the text into a sequence of tokens, which can be words or sub-words. 

We then removed all stop words. Stop words in natural language are the most common words which do not convey significant meaning; for example, in English these are words such as `the', `for', `at', `to', etc. English and French dictionaries were used to identify and remove the stop words. We also perform lemmatization of the tokens, which refers to converting a token to its `base' or `dictionary' form. This process includes, for instance, plural to singular conversion, as well as word tense normalization. Unlike stemming (another popular approach which simply removes affixes), lemmatization takes into account a broader context, including the surrounding and adjacent clauses.
Finally, we employ N-grams, which are continuous sequences of tokens in a document. Bigrams (2) and trigrams (3) were used in our analysis. 

For the final step, Latent Dirichlet Allocation (LDA) was used as a statistical model for the abstracts. LDA models every document as a mixture of topics, and every topic as a mixture of words. Crucially, it is possible to simultaneously estimate both parameters: finding a combination of words associated with each topic, as well as determining a combination of topics that describe each document (in this case abstracts). The LDA model was built with 30 different topics, where each topic is a combination of keywords and each keyword contributes a certain weight to the topic.
After training, we obtained the following evaluation for the LDA model:

\hfill \break
\centerline{Perplexity:  -15.231877825584922}
\centerline{Coherence Score:  0.4339549814094681}\\

Perplexity measures how well a probability distribution predicts a sample; it thus indicates how well the topic model fits the data. Coherence, on the other hand, measures the degree of semantic similarity between words in a topic, in order to measure how `meaningful' each topic is. %And by combining them together with coherence, you can get a convenient measure of how good this model is. 
The lower the perplexity, the better the model, and vice versa for coherence. While perplexity is an important measure of fit, we are more interested here in obtaining a conceptually coherent/consistent topic model. Therefore, in this analysis, we primarily consider the coherence score.
%But too low values of perplexity may also not make sense, as the conceptual value of the model will not be increased. 

According to non-optimised LDA analysis, a vast proportion of the retrieved articles are in the field of research on cancer and mental disorders. This is indicated by word markers such as `death', `cognitive', `brain', `emotion' and `behaviour' (see Appendix A A4-A8). 
On the charts, circles represent the topics. The larger the circle, the more common this theme. A good topic model should have large non-overlapping circles scattered throughout the diagram. In turn, a model with too many topics will often have many small circles that can overlap and be in the same area of the chart.

To optimize the model, hyperparameter tuning was performed. For this, the coherence score was calculated for 10 topics, for a range of values of alpha and beta hyperparameters. The maximal coherence score was obtained using 10 topics, alpha = 0.01,  and beta=0.91. After the selection of hyperparameters, the accuracy of the model was:

\hfill \break
    \centerline{Perplexity:  -8.210963741472142} 
    \centerline{Coherence Score:  0.643855950055103}\\

The model has become more coherent, but less sensitive. In Figure A7 in Appendix A, we see that the top-30 most salient items include tokens such as `brain', `cognitive', `death', `neural', `intervention', `memory' and `child'. According to the model, a possible interpretation is that the interventions are probabilistically relevant to 37.5\% of the topics in which the study was conducted. In the distance map, the first bubble includes generic terms that can be found in almost all medical works. The behaviour token is relevant to three large clusters; it can be ignored. The second large cluster includes tokens such as `intervention', `death', `treatment', and `child'. This may be due to the fact that research on interventions is mainly conducted on fatal diseases that children suffer from. The third large cluster of tokens relates to diseases associated with a decline of brain function. This can be observed if the relevance metric ($\lambda$) is 0.6. The fourth and fifth clusters are language clusters, such as Spanish and French. Other clusters represent a small percentage of the total and are mostly devoid of particular semantics.

In Appendix A, Figure A2 shows the search results for all three databases. From ScienceDirect, in the Computer Science field 257 papers were downloaded (out of 258), in the Medicine and Dentistry fields 609 papers were downloaded (out of 632), in the Neuroscience field 342 papers were downloaded  (out of 382), and in the Psychology field 271 papers were downloaded  (out of 305). In total 1479 papers was downloaded and after duplicate removal 1093 papers were retained. Our overall finding is that the bulk of the analyzed work is in the areas of cancer, respiratory diseases and brain research. Interventions are likely to be found in studies that investigate disease deaths (e.g., cancer and COVID-19). A search was carried out for relevant data in the articles. Due to missing data, only 34 articles were included in this report.

\section*{Review of Relevant Articles}
\addcontentsline{toc}{section}{Review of Relevant Articles}
Due to variability in presentation of diseases, it can be difficult for doctors to prescribe optimal treatment plans. More accurate predictive models can be beneficial to provide personalised treatment and determine effective interventions. At the same time, it is important that any interventions we apply are safe; a medical error in any specific intervention could be fatal. This may be due to the individual reactions of the patient's body to drugs or on the part of the doctor, for example, loss of concentration. From a practical perspective, rigorously testing all potential interventions/treatments for all types of patients in long clinical trials is not always feasible as valuable time may be lost, and the result may not be positive. As a result, doctors often have to rely only on their experience, and the wrong choice of intervention may aggravate the course of the disease. In this review, we are interested in applications of causal inference to help solve the `treatment effect' problem. In this context, one aims to choose an intervention which has a positive causal effect on the outcome variable, which will help to determine the most effective treatment plan. The effect of an intervention can be formalized as the difference between two potential outcomes for an individual: one where they receive the intervention, and one where they do not. As a result, сounterfactuals are the basis for causal conclusions, giving answers to a number of questions \cite{Chaieb2021}. For example:

\begin{enumerate}
    \item will the intervention work? 
    \item why did it work? 
    \item what combination of interventions are safe/can reduce the risks? 
    \item how effective is the intervention?
\end{enumerate}

Unfortunately, in many articles, the concept of intervention is extremely vague. In particular, as mentioned previously, in medical articles interventions are usually understood as operational actions rather than part of a model. Also, in more than 90\% of cases, data or models are not available. Thus, we selected all papers with medical data available for review. A brief summary of the relevant articles now follows.

Several papers %[2-11] \marta{add citations to individual papers} 
demonstrate applications of Bayesian network analysis in cancer research.

In \cite{sesen_bayesian_2013}, the authors build a Bayesian network model using the English Lung Cancer Database (LUCADA) and apply causal interventions. The results show that the survival-maximizing model only works correctly in 29\% of cases. This is due to the lack of some of the information necessary for effective interventions; in the database, the constructed model is not able to distinguish between patients for whom surgery is indicated and those for whom it is not suitable. The model and data are not available. The authors continue their work in \cite{sesen_lung_2014}, where a clinical decision support application for lung cancer care (LCA) was developed. The goal was to help lung cancer experts make decisions about treatment choices in MDT (Multidisciplinary Team) meetings. The system is not available via the link in the article (the login page is stored in the webarchive).

The impact of obesity in early-stage breast cancer survivors on health behaviors was studied in \cite{xu_modeling_2018}. Clinical data from 333 overweight or obese postmenopausal women who survived breast cancer were used to build a Bayesian network. The authors combine factors to determine possible interventions. For example, based on the network they developed, they conclude that older age, more sleep disturbances, and higher BMI are associated with lower physical activity. Data is available in the article, but presumably, after the anonymization of the personal data, some parameters were removed. Because of this, the dataset is incomplete. It may be possible to synthesize the missing data based on what is already known.

Factors and causal relationships influencing the choice of place of death in Swiss cancer patients were studied in \cite{kern_impact_nodate}. One of the goals of the project was to explore the possibility of interventions provided by healthcare professionals to facilitate EOL (end of life) at home. The model was built on 116 adult patients who died from cancer between 2015 and 2016 in southern Switzerland. The model can be downloaded from the supplementary materials. Data is available on request.

Using data from the National Lung Screening Trial (NLST), a model is built in \cite{petousis_prediction_2016} to study the impact of interventions on the survival of patients with lung cancer. By interventions, the authors mean low-dose computed tomography (LDCT) irradiation to reduce patient mortality. The authors constructed their dynamic Bayesian network (DBN) using 10-fold cross-validation. The models were evaluated based on the probabilistic variable of the biopsy, taking into account all previous and current data for each of the three intervention points in the NLST study. Data is available upon request.

For the diagnosis of breast cancer, researchers used a Bayesian network \cite{liu_quantitative_2018} built using the K2 algorithm. The data is collected from The First Affiliated Hospital of Fujian Medical University, China and the Breast Cancer Wisconsin Dataset (BCWD) of the UCI machine learning repository. Naive Bayes (NB), Bayesian network (BN), ID3, J48 and NBTree classifiers are used for performance analysis. The UCI dataset is available for download.

In \cite{kourou_cancer_2020}, the authors (using transcriptomic datasets) build a DBN model used for cancer classification. They used three datasets for a DBN-based approach, which is able to model time series gene expression data for classification purposes. The datasets are the pancreatic cancer dataset (GSE14426), colon cancer dataset (GSE37182), and breast cancer dataset (GSE5462). Data is available through GEOquery package.

Analyzing multiple primary cancers is a challenge. In the study of \cite{wang_survivability_2020}, the authors build a BN model to describe the occurrence of two primary cancers to predict five-year patient survival. They compare the performance of models with BPNN, Logistic Regression (LR), Support Vector Machine (SVM) and Naive Bayes (NB). Resulting conditional probabilities and the model are available in the paper.

Preoperative identification of patients at risk for lymph node metastasis (LNM) is challenging in endometrial cancer \cite{reijnen_preoperative_2020}. The authors develop a BN model to predict LNM and outcome in endometrial cancer patients. They constructed the dataset from a retrospective multicenter development cohort from 10 centers across Europe. The model predicts 1, 3 and 5-year survival and was tested on 2 external cohorts from the Netherlands and from Norway. The trained model is available.

Breast cancer is the most common cancer in women. Clinicians face difficult decisions in many aspects of breast cancer treatment. In  \cite{jiang_clinical_2019}, the authors developed a Bayesian network called Causal Modeling with Internal Layers (CAMIL) for preventing breast cancer metastasis. A decision support system (DSS) based on this model, called DPAC, was developed and tested by 5-fold cross-validation. The system can recommend related treatments for patients based on not only mutation profile but also their histopathology and clinical parameters. A dataset called LSDS-5YDM is available.

Due to the emergence of COVID-19 disease, there is a need for Contact Tracing Apps (CTA). But there have been concerns about the privacy of such applications. In \cite{fenton_privacy-preserving_2020}, the authors propose a model based on the user's decisions about the confidential data provided to them. Based on the data, the model predicts the likelihood of the presence of asymptotic, mild or severe COVID-19. The authors performed a sensitivity analysis for the `eventual COVID19 status' node. It shows which yet unobserved factors have the greatest impact on the target node. The full BN model is available.

In \cite{akinnuwesi_application_2021}, researchers studied what model is best for classifying COVID-19 given overall symptoms. They conducted experiments with Logistic Regression (LR), Support Vector Machine (SVM), Naive Bayes (NB), Decision Tree (DT), Multilayer Perceptron (MLP), Fuzzy Cognitive Map (FCM) and Deep Neural Network (DNN) algorithms. It was observed that the model depends on the chosen resampling technique. 
The dataset with numerical variables is available.

In \cite{wang_application_2021}, the authors built a model for the spread of hyperlipidemia and related factors in Shanxi Province. To construct the BN structure, the authors compared several algorithms, in particular MMPC-Tabu, Fast.iamb-Tabu and Inter.iamb-Tabu. To build the model, the authors use the Inter.iamb-Tabu hybrid algorithm. According to the model, it can be said that gender, BMI, and physical activity are directly related to hyperlipidemia. The accompanying materials provide the source code; the data is available on request.

Using logistic regression with univariate and multivariate models, the authors investigated the main correlates of cirrhosis complicated by encephalopathy (HE) \cite{zhang_application_2019}. Infection, electrolyte disorder and hepatorenal syndrome have been found to have a strong influence on the occurrence of HE. For evaluation, the authors used AUC (area under the receiver operating characteristic curve). For the reasoning model, they used deviations from the normal state of a person.
The trained model is available in Appendix C in Figure C.2.

In \cite{loghmanpour_new_2015}, the authors predict mortality in patients after implantation of left ventricular assist devices. In addition to other factors influencing mortality, the model takes into account interventions in the last 48 hours. They compare the results of the BN model and HMRS. The model is available. Data is available upon request from the Interagency Registry for Mechanically Assisted Circulatory Support (INTERMACS) database.

Osteomyelitis (OM) is a bone infection that occurs more often in younger children. In \cite{wu_predicting_2020}, the authors used a BN to select the most appropriate antibiotic therapy based on data on the OM pathogen. The authors combine variance-based sensitivity analysis and expert surveys about certainty to identify the greatest influence on pathogen nodes. For validation the AUC was used. Data is not available.

Decisions in transplant care often involve trade-offs between potential benefit and potential harm of treatment or intervention. In  \cite{neapolitan_primer_2016}, the authors build a model for making kidney transplant decisions. A decision-making method based on Kidney Donor Risk Index (KDRI) was developed. The model predicts the probability of death over the next 3 years. Data is not available.

Assessing and managing the risk of violence is considered a critical component for decision making in medium secure services \cite{constantinou_causal_2015}. The researchers build a model in forensic psychiatry  using a dataset referred to VoRAMSS. In order to find out the impact on the output node, sensitivity analysis is performed. The model is compared with other regression models trained on the same dataset. Data not available.

To assess and manage the risk of future violent re-offending for released prisoners with mental illness, a DSVM-P model was developed in \cite{constantinou_risk_2015}. The authors explore the issue of reducing the risk of future re-offending to an acceptable level by introducing interventions. Using sensitivity analysis applied to the interventions, the authors observe the most active symptoms. Data is not available.

A causally-based decision support tool for the management of violence risk among released prisoners using Bayesian networks was created in \cite{Coid2016-of}. The risk analysis is managed and analysed through interventions on variables such as alcohol, drugs and psychiatric interventions. The model was compared with the results from HRC-20, VRAG and PCL-R risk assessment instruments. Data is not available.
The process was replicated in \cite{Coid2016-yz}. The authors develop a risk management tool for discharged forensic prisoners using the same methodology. The model was compared with the results from HRC-20, SAPROF and PANSS risk assessment instruments. The data is not available.

Autism, or autism spectrum disorder (ASD), occurs due to developmental disorders caused by differences in the brain. For predicting autistic spectrum disorder, it is possible to use fuzzy cognitive maps (FCM) \cite{kannappan_analyzing_2011}. The authors used a non-linear hebbian learning for training model. In the study, experts, based on the constructed model, determine the boundaries for the acceptable probabilities of having autism. They define: definite autism, probable autism and no autism. The data is available.

A Bayesian network decision model was proposed for supporting the diagnosis of dementia, Alzheimer’s disease (AD), and mild cognitive impairment (MCI) \cite{seixas_bayesian_2014}. In the paper the list of CDSS (clinical decision support system) was provided. The authors perform a sensitivity analysis for parameters like dementia, AD and mild cognitive impairment. The performance was tested by AUC, F1, MSE, MXE scores. The dataset based on data from the Duke University Medical Center and the Center for Alzheimer's Disease and Related Disorders can be purchased.

Chronic obstructive pulmonary disease (COPD) is a common, preventable and treatable disease, with a worldwide prevalence of 10.1\% in people aged 40 years or older \cite{Jarhyan2022}. Using a Bayesian model, the authors analyse the impact of critical variables on the risk of rehospitalization in patients with COPD \cite{lee_reducing_2018}. The authors argue that it is possible to reduce these risks with the help of personalised interventions. The accuracy of the model is determined in comparison with other machine learning methods. The dataset was constructed based on St. Mary's Hospital data.

In \cite{dag_probabilistic_2016} and \cite{topuz_predicting_2018}, the authors used machine learning (ML) techniques and built transplant survival models. For predicting the outcome of a heart transplant, the authors built a framework based on a BN model that can identify patient-specific survival risk scores and the interactions between the explanatory variables \cite{dag_probabilistic_2016}. For testing performance of the model six evaluation metrics were used. The dataset was provided by United Network for Organ Sharing (UNOS). The dataset and software are available for download.
Of the 119,873 organ transplants worldwide, 67\% are kidney transplants, and 59\% of those are from deceased donors, according to the World Health Organization \cite{who}. A three-step Bayesian risk model for predicting graft survival in kidney transplantation was proposed in \cite{topuz_predicting_2018}. The UNOS data is used for building the model. The authors used several techniques, such as ANN (artificial neural network) and SVM (support vector machine), and sensitivity analysis for selecting features for the final dataset. For performance measure of the accuracy, sensitivity, specificity, f-measure, and g-mean metrics were used. %Risk levels of the model are as described in Section 3. 
Data is not available.

Approximately 537 million adults (20-79 years) are living with diabetes. The total number of people living with diabetes is projected to rise to 643 million by 2030 and 783 million by 2045 \cite{idf_da}.  For the prevention of complications of type 2 diabetes mellitus (T2DM), the authors propose a model that predicts six complications of T2DM \cite{liu_analysis_2020}. The model was built using the Tabu-search and Bootstrap algorithms. Data was taken from the National Health Clinical Center for the period 1st January 2009 to 31st December 2009. The authors compare the performance between BN, BN-wopi, NB, RF and C5.0 models using AUC, 95\%CI, sensitivity and specificity metrics. The article also provides a comparative performance table of models from related studies by AUC score. Dataset is available on request.

In contrast, \cite{marini_dynamic_2015} proposes to predict the complications associated with type 1 diabetes. The paper presents two types of models using two different approaches (DDO-DBN, EI-DBN) in the design of the structure of the Dynamic Bayesian Network. The difference is that the second network is built on the basis of knowledge from the medical literature. As a result of the analysis, the authors determine threshold values for continuous variables. The authors used DCCT and EDIC data sets for building the model.

In \cite{shen_cbn_2018}, the possibility of automatic construction of the Bayesian network topology and ontology from electronic medical records using the K2 algorithm was studied. The dataset was built on the basis of 10,000 anonymized patient records.
%, subsequently adopted the odds ratio (OR value) calculation. 
The classification performance of the constructed CBN is compared with other inference models such as Naive Bayes, Basic, Random-node-input, and Frequency-based Bayesian networks. The data is available.

Infectious diseases are the world's greatest killers, accounting for more than 13 million deaths annually among children and young adults alone \cite{inf_tf}.  Patients often self-medicate to treat infectious diseases. In such cases, a system is required that can help in the choice of antibiotics. In \cite{shen_ontology-driven_2018}, the authors propose a system called IDDAP based on ontologies for infectious disease diagnosis and antibiotic therapy. The ontology system was built by combining existing ontologies. Performance analysis was carried out using ROC, and the system was tested on different ontologies. The ontology is available; the software has been removed from Github.

Non-communicable diseases (NCDs) kill  41 million people each year, equivalent to 71\% of all deaths globally \cite{cghe}.  Prevention of these diseases can help in their development. In \cite{pittoli_intelligent_2018}, the authors developed a system based on Bayesian networks. The system determines the impact of user interventions on the main risk factors. The task was to predict coronary heart disease. The desired health indications are also indicated, i.e. upper and lower limits of the normal state without complications. A number of datasets were used in this work; Pima Indians Diabetes Data Set is available on Kaggle.
 
While much attention has been paid to the problem of evaluating the effectiveness of therapy using observational data, little work has been done on evaluating the treatment effect of interventions \cite{chaieb_personal_2022}. In the paper, the authors propose a model that studies reducing the number of falls in elderly patients through interventions in the model. Data from 1810 patients at Lille University Hospital (France) are used to build models. Experiments were carried out with three prediction models (BN, SVM, and DT); BN structure and modalities are available. 
%The dataset is shortest.

There are a number of tasks in which it is required to make timely diagnostic and managerial decisions. One such challenge is diagnosing infections in children in the emergency department. Based on expert opinion and knowledge in the area of expertise, \cite{ramsay_urinary_2022} developed a Bayesian model to predict the results of causative pathogens. The paper provides three clinical scenarios to support decision making by physicians. The model is available.

In \cite{pan_application_2019}, the authors identify factors associated with hypertension. To achieve this, they build a Bayesian network based on Tabu search. The study participants were selected only from cities in Shanxi province, which does not allow generalisation of the results to a wider population. For evaluation, the authors used accuracy, TPR, FPR, precision, recall and F-measure metrics using the Weka software. The model has parameters such as smoking or drinking that can be used as interventions. Dataset and model are available. 

Finally, we remark that in Appendix B (see Tables 1 and 2) we summarise information on the availability of medical data. Data was collected through a systematic review of publications. In most papers, data from the UCI and Kaggle repositories are used. According to the analysis of papers, several articles were identified that met the criteria stated above. The case of interventions for fall prevention in elderly patients \cite{chaieb_personal_2022} was chosen as the best suited to further analysis.

\section*{Interventions and Disease Management: Risk Boundary Selection}
\addcontentsline{toc}{section}{Interventions and Diseases Management: Risk Boundary Selection}
In the medical field, prediction of the risk of disease requires establishing a statistical model of risk factors and diseases, and prediction of the probability of disease according to levels of multiple risk factors \cite{Zhang2016}.

Usually, the reasoning about the risk boundaries is based on deviations from the normal state, as for example in \cite{sesen_bayesian_2013}, \cite{zhang_application_2019}, \cite{kannappan_analyzing_2011}, \cite{Daudn2012}. In the medical literature, there are also works which establish some limits/bounds on the observed variables that indicate the normal state of a patient, with values outside the limits indicating the presence of the disease. This section provides several examples of such works.

Some articles have outlined specific risk boundaries \cite{reijnen_preoperative_2020}, \cite{jiang_clinical_2019},\cite{dag_probabilistic_2016}, \cite{topuz_predicting_2018}, \cite{pittoli_intelligent_2018}, which can be used to evaluate interventions and, therefore, to recommend more effective personal therapy.

In \cite{topuz_predicting_2018}, the authors set boundaries for the response to transplantation. There are three risk groups defined: low, medium and high risk. Depending on which risk group a patient falls in, it is possible to choose personalized treatment more accurately, and, more importantly, to prevent a patient from falling into a high-risk category in advance, by determining the influence of certain factors on risk.

In \cite{pittoli_intelligent_2018} interval boundaries are indicated for CAD risk factors. Also in this work, the authors provide a comparative analysis of similar approaches, comparing them in terms of parameters of risk factors analysis and context awareness. Based on the model, the authors developed an application that provides recommendations on factors that can exacerbate or alleviate the course of the disease. To do this, they created several scenarios and tested the system on two datasets.

An alternative to using boundaries is finding the most optimal solution using influence diagrams, i.e., a Bayesian network augmented by utility nodes (UN).

The authors of \cite{neapolitan_primer_2016} suggest using the Kidney Donor Risk Index (KDRI) to evaluate complex transplant decisions. The metric assesses the risk of transplant rejection compared to a 40-year-old healthy donor. The model was built in Netica using UN. The purpose of using utility nodes is to maximize the expected value while looking for the most optimal solution. %UN variable tables are given in the article.

Regarding acceptable levels of numerical variables, \cite{Daudn2012} presents an algorithm for the treatment of diseases associated with psoriasis. For example, for the treatment of metabolic syndrome, levels are set for HDL, AHT, BP, and TG. Using the reasoning model of this work as an example, experts can build BN models, determine acceptable levels of variables, and design effective interventions.

In \cite{Buvat2013}, algorithms for the diagnosis of testosterone deficiency and follow up of testosterone therapy are developed. From analysis of the literature, the authors identify the main symptoms, signs, and conditions indicative of testosterone deficiency. The levels of physical indicators
 %, and algorithms of influences that can be applied when achieving them 
 are also given. 
 The information provided in the article can be used to design intervention-based CDSS.

In an infection control program studying the implementation of preventive measures, the use of data and surveillance audits is central. To improve these practices, a list of interventions can be compiled based on CDC recommendations and practice monitoring \cite{kdigo}. This work provides recommendation guidelines for the prevention, diagnosis, and management of Hepatitis C (HCV) in patients with chronic kidney disease (CKD). The paper presents treatment schemes and risk grade for CKD, which can be used as a definition of acceptable boundaries of the target node and effective interventions.

Interventions can also be derived from literature analysis, as was done in \cite{Watt2017}. As a result of the analysis, the authors proposed pharmacological and non-pharmacological interventions for BPSD (dementia symptoms).

Based on the works presented in this section (and similar other works), it is possible to develop more effective systems for controlling risk levels based on interventions, which can be the next step in building effective CDSS systems. \hfill \break

\section*{Review Summary and Model Selection}
\addcontentsline{toc}{section}{Review Summary and Model Selection}
In most of the analyzed papers, machine learning methods like support vector machines (SVM), Bootstrap Forest (BF), Bayesian belief network (BBN), and artificial neural networks (ANN) were used. When BN models are used, they may be part of a larger pipeline \cite{dag_probabilistic_2016} if the data was not initially prepared/in the right format to work with BNs. AUC is most commonly used for accuracy evaluation. Bayesian network models are mostly built on the basis of expert opinions \cite{ramsay_urinary_2022}, but sometimes they are also built using the K2 \cite{shen_cbn_2018} or TabuSearch \cite{wang_application_2021} algorithms, then checked or supplemented by experts \cite{chaieb_personal_2022}.
These authors use their own data derived from close cooperation with medical institutions (Table B1 of Appendix B)  or data from repositories (Table B2) or medical databases (Table B3). Also, in some works, data from multiple sources can be combined if incomplete \cite{pittoli_intelligent_2018}. %or checking model for few deceases .
By interventions, the authors mostly refer to medical interventions that are already accounted for in the Bayesian network \cite{shen_ontology-driven_2018}. In some works, sensitivity analysis is carried out to determine the impact of various factors on the target node \cite{fenton_privacy-preserving_2020}, \cite{dag_probabilistic_2016} according to some machine learning classifier. Similarly, it is possible to use `what-if analysis' in Bayesian networks to discern the effect of a given variable on the target by conditioning on the value of that variable \cite{topuz_predicting_2018}. This could be used as a heuristic for selecting promising interventions to study.
%In turn, sensitivity analysis shows the main causal factors for the target node.
The main evaluation indices of a BN model are true positive rate (TPR), true negative rate (TNR), recall, and precision. Sensitivity (TPR) indicates the proportion of positive classes correctly predicted and the ability of the BN to recognize positive classes. Specificity (TNR) represents the proportion of correctly predicted negative classes and measures the ability of the BN to recognize negative classes \cite{pan_application_2019}.

In Table B1 (Appendix B), the found models are highlighted in green, and the proposed model for our experimental study of interventions is highlighted in blue (a shorter dataset was attached to the work). Appendix C contains the constructed models based on the available data in the papers.

\section*{Experimental Evaluation}
\addcontentsline{toc}{section}{Experimental Evaluation}
This section of the report summarises our experiments on selected models using the software tool developed in~\cite{Wang2021}. Our workflow was primarily based upon the framework/pipeline illustrated in Appendix D for Bayesian network modelling, based on \cite{Kyrimi2021}. In the following experiments, since we already have access to the Bayesian network from the literature, we skipped the first few steps. For interventional modelling, we make use of the IntRob (interventional robustness) software \cite{Wang2021}, based on a recently proposed approach to providing guarantees on the effects of combinations of interventions. As we will see, this is important for providing safety guarantees with respect to different treatment (intervention) policies.

We identified the Bayesian network models from \cite{reijnen_preoperative_2020, jiang_clinical_2019, chaieb_personal_2022} as particularly suitable for interventional analysis; in particular, we conduct experiments on the basis of the model from paper \cite{reijnen_preoperative_2020}. 
In this paper, the authors attempt to identify patients at risk for lymph node metastasis (LNM) in endometrial cancer. The Bayesian network model can be used to predict lymph node metastasis, as well as outcomes such as patient survival up to 5 years and recurrence of disease, given some (incomplete) evidence/information about the patient. 
%Interventions can be recurrence and adjuvant therapy.

An oncologist treating a cancer patient employs the following procedure. In the first stage, a diagnosis is made and it is determined how much the tumor has spread and whether the patient is operable. In the second stage, if the patient is operable, then they apply radiation therapy targeting the tumor and nearby tissues (this depends on many factors). After irradiation, the patient is surgically operated on, with  irradiated and nearby tissue removed. Finally, chemotherapy is prescribed to prevent spread of tumour cells into lymph nodes or blood.

In their work, the authors focus on predicting lymph node metastasis, which is one of the most important prognostic factors for patient outcomes, and which can be significantly influenced by adjuvant therapy: it is thus of significant practical interest to be able to predict LNM in order to inform therapy decisions. We begin by performing feature selection for the classifier based on their impact on the LNM probability in the Bayesian network. In Table 1, we show the probability of LNM and probability of LVSI (lymph-vascular space invasion), for a range of different evidence values. 
%the probability table has been refined and an LVSI (lymph-vascular space invasion) node has been added to select nodes for the classifier. T
The green cells are updated values and the blue one was chosen for classification. \\

%\newpage

{\bf Table 1}. Probability estimates for lymph node metastasis and LVSI given different evidence \cite{reijnen_preoperative_2020} 
        \begin{longtable}{C{4.5cm}  C{3cm}  C{3cm}  C{1.5cm} }
            \toprule
            
            \makecell{Evidence provided \\ to the Bayesian network} & \makecell{Modalities}  & \makecell{Lymph node \\ metastasis} & \makecell{LVSI} \\ 
            \midrule
             No evidence	 & &	8.6	& 16 \\
            \midrule

            \multirow{1}*{Preoperative grade}
                & \multicolumn{1}{c }{1} 	 & \multicolumn{1}{c }{4.7}  & \multicolumn{1}{c }{10.6} \\ 
                & \multicolumn{1}{c }{2} 	 & \multicolumn{1}{c }{\colorbox{green}{8.2}}  & \multicolumn{1}{c }{14.6} \\
                & \multicolumn{1}{c }{3} 	 & \multicolumn{1}{c }{\colorbox{green}{20.7}} & \multicolumn{1}{c }{34.4} \\
                \midrule
			\multirow{1}*{Preoperative grade, L1CAM}
                & \multicolumn{1}{c }{1,negative} & \multicolumn{1}{c }{\colorbox{green}{3.9}} & \multicolumn{1}{c }{9.9} \\ 
                & \multicolumn{1}{c }{1,positive} & \multicolumn{1}{c }{\colorbox{green}{17.7}} & \multicolumn{1}{c }{22} \\ 
                & \multicolumn{1}{c }{2,negative} & \multicolumn{1}{c }{6.6} & \multicolumn{1}{c }{13.1} \\ 
                & \multicolumn{1}{c }{2,positive} & \multicolumn{1}{c }{28.2} & \multicolumn{1}{c }{28.1} \\ 
                & \multicolumn{1}{c }{3,negative} & \multicolumn{1}{c }{\colorbox{green}{17.4}} & \multicolumn{1}{c }{31.8} \\
                & \multicolumn{1}{c }{3,positive} &  \multicolumn{1}{c }{\colorbox{green}{30.4}} & \multicolumn{1}{c }{41.8} \\ 
                \midrule
            \multirow{1}*{Preoperative grade, ***PM}
                & \multicolumn{1}{c }{1,favorable*} & \multicolumn{1}{c }{2.8} & \multicolumn{1}{c }{8.9} \\ 
                & \multicolumn{1}{c }{\colorbox{blue!30}{1,unfavorable**}} & \multicolumn{1}{c }{\colorbox{green}{35.8}} & \multicolumn{1}{c }{45} \\ 
                & \multicolumn{1}{c }{2,favorable} & \multicolumn{1}{c }{\colorbox{green}{5.1}} & \multicolumn{1}{c }{11.6} \\ 
                & \multicolumn{1}{c }{2,unfavorable} & \multicolumn{1}{c }{\colorbox{green}{36.2}} & \multicolumn{1}{c }{43.4} \\ 
                & \multicolumn{1}{c }{3,favorable} & \multicolumn{1}{c }{\colorbox{green}{16.9}} & \multicolumn{1}{c }{29.3} \\
                & \multicolumn{1}{c }{3,unfavorable} &  \multicolumn{1}{c }{37.1} & \multicolumn{1}{c }{45.4} \\ 
                \midrule
            \multirow{1}*{Preoperative grade, }{Ca-125}
                & \multicolumn{1}{c }{1,normal} & \multicolumn{1}{c }{1.5} & \multicolumn{1}{c }{8.9} \\ 
                & \multicolumn{1}{c }{1,elevated} & \multicolumn{1}{c }{\colorbox{green}{22.8}} & \multicolumn{1}{c }{20.3} \\ 
                & \multicolumn{1}{c }{2,normal} & \multicolumn{1}{c }{\colorbox{green}{2.8}} & \multicolumn{1}{c }{11.8} \\ 
                & \multicolumn{1}{c }{\colorbox{blue!30}{2,elevated}} & \multicolumn{1}{c }{\colorbox{green}{34.7}} & \multicolumn{1}{c }{28.2} \\ 
                & \multicolumn{1}{c }{\colorbox{blue!30}{3,normal}} & \multicolumn{1}{c }{\colorbox{green}{7.7}} & \multicolumn{1}{c }{28.5} \\
                & \multicolumn{1}{c }{3,elevated} &  \multicolumn{1}{c }{\colorbox{green}{60.9}} & \multicolumn{1}{c }{52.2} \\ 
                \bottomrule

    \end{longtable}
    \noindent {\small *Favorable: all IHC stainings were normal (ER, PR positive, L1CAM negative, p53 wildtype). \\ 
    **Unfavorable (ER, PR negative, L1CAM positive, p53 mutant). Ca-125, cancer antigen 125; ER, estrogen receptor; L1CAM, L1 cell adhesion molecule; PR, progesterone receptor. \\
    ***PM - molecular profile.} \\

    In Table 2, we show a classification of risk groups for LNM. \\

    {\bf Table 2}. LNM risk table \cite{reijnen_preoperative_2020}
    \begin{longtable}{ C{3cm}  C{3cm} }
            \toprule
            
            \makecell{Predicted \\ probability} &	\makecell{Risk group} \\
            \midrule
                 $<$ 1\% &	Very low \\
      
                1\%-5\%	& Low \\
             
                6\%-15\%  &	Intermediate \\
    
                16\%-25\%  & High-intermediate \\
        
                $>$ 25\%  & High \\

            \bottomrule
            \endhead
    \end{longtable}
    
    According to Table 1, the following variables were selected as features for the classifier: ER, PR, L1CAM, p53, preoperative grade, CA125. We used a threshold of 0.05 for the prediction of LNM, in accordance with the risk profiles in Table 2. We use the BNC\_SDD software to obtain a logical representation of the resulting Bayesian network classifier (BNC), which we use for further analysis in the IntRob software.

    \begin{figure}[htb]
       \centering
        \includegraphics[keepaspectratio]{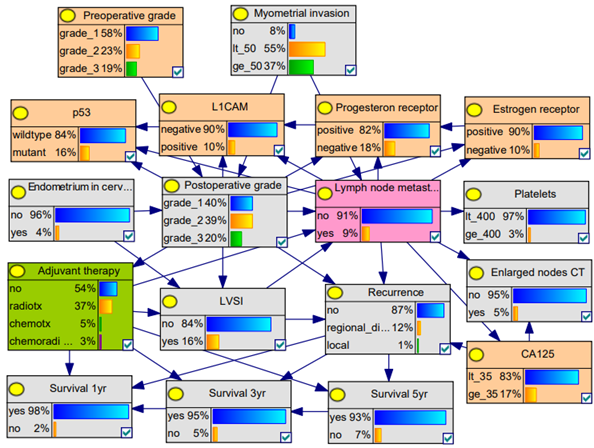}
         \caption{Fig. 1 The chosen Bayesian network. The target node (LNM) is shown in pink. Yellow nodes was chosen as features for classification, while green nodes represent intervention targets.}
          \label{fig:e1}
    \end{figure}

    \begin{figure}[htb]
       \centering
        \includegraphics[keepaspectratio]{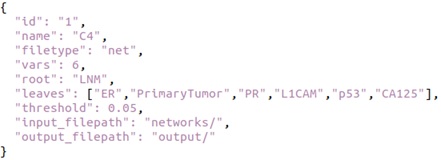}
         \caption{Fig.2 JSON file %for. odd % 
         for LNM classifier}
          \label{fig:e2}
    \end{figure}
    
    For our first experiment, we considered the probability of LNM under interventions on adjuvant therapy. Such an intervention represents a change in treatment policy; it can be seen, in particular, that adjuvant therapy has a causal influence on LNM from the Bayesian network graph. In particular, we asked the following question: what are the potential consequences of instituting a bad treatment policy? To analyze this, we used the IntRob software to compute an upper bound on the probability of lymph node metastasis, under all possible interventions (treatment policies). In Figure 5, we see that the worst-case policy may result in 60\% of patients suffering metastasis, up from 9\% under the current policy. This shows that it is vital to choose the right adjuvant therapy for each patient depending on their characteristics, and failing to do so can be unsafe.
    In Figure 6, we conducted another test to see how different interventions on adjuvant therapy affect the false negative rate of the trained classifier; that is, whether this might result in the classifier mispredicting patients as not at risk of LNM when they actually are. Interestingly, the classifier is surprisingly robust, with a false negative probability of only 0.001. This can possibly be attributed to the fact that the classifier uses indicators causally downstream from LNM (p53, L1CAM, etc.), which provide sufficient information to correctly predict LNM, even with the shift observed above in the LNM distribution.

    The results of the working of IntRob software can be seen in Figures 3-5.
    \begin{figure}[htb]
       \centering
        \includegraphics[width=1\linewidth, angle=0]{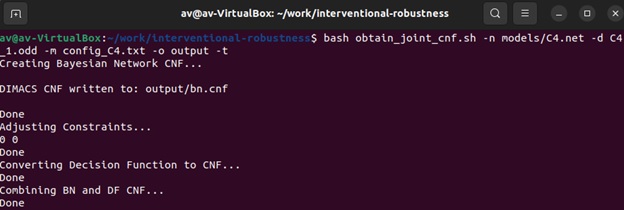}
         \caption{Fig. 3 First four steps of the IntRob algorithm}
          \label{fig:e3}
    \end{figure} 
    
    \begin{figure}[htb]

       \centering
        \includegraphics[width=1\linewidth, angle=0]{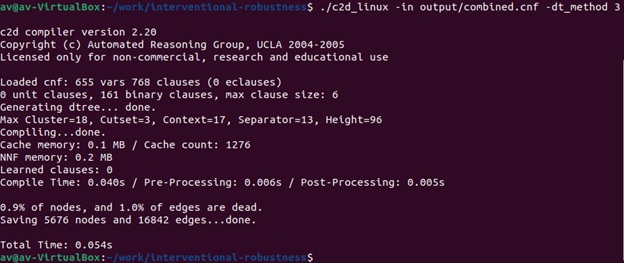}
         \caption{Fig. 4 c2d compiler output}
          \label{fig:e4}
    \end{figure}
    \begin{figure}[hbt]

       \centering
        \includegraphics[width=1\linewidth, angle=0]{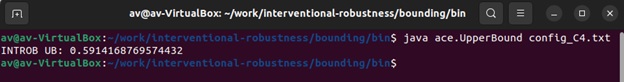}
         \caption{Fig. 5 Upper bound on LNM probability under worst-case intervention}
          \label{fig:e5}
    \end{figure}
    \begin{figure}[htb]
       \centering
        \includegraphics[width=1\linewidth, angle=0]{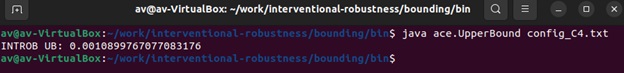}
         \caption{Fig. 6 Upper bound on false negative probability}
          \label{fig:e6}
    
    \end{figure}
   \hfill \break
   \begin{figure}[htb]
   For the second experiment, we study the probability of recurrence, again under interventions to the adjuvant therapy. Applying the IntRob software to derive the worst-case probability of recurrence over all possible interventions, we find in Figure 8 that the probability of (regional-distant) recurrence can reach as high as 38\%, once again demonstrating the risks of inappropriate treatment. 
   %The ordered decision diagram was made for the followings nodes: p53, PrimaryTumor, CA125, ER, PR, L1CAM.
   \end{figure}
   \begin{figure}[htb]
       \centering
        \includegraphics[width=1\linewidth, angle=0]{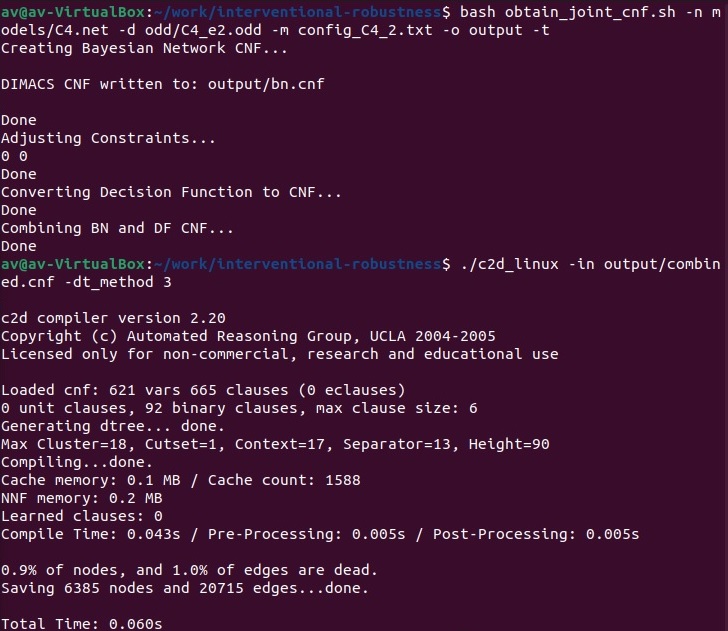}
         \caption{Fig. 7 IntRob software for second experiment}
          \label{fig:e7}
    \end{figure}
    \begin{figure}[htb]
       \centering
        \includegraphics[width=1\linewidth, angle=0]{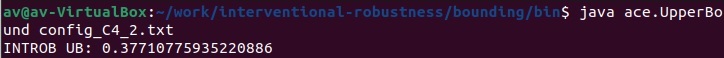}
         \caption{Fig. 8 Upper bound on recurrence probability under worst-case intervention}
          \label{fig:e8}
    \end{figure}
    
    \begin{figure}[htb]
    Finally, for the last experiment, we considered 5-year survival (`Survival 5yr'). We performed a sensitivity analysis in Figure 9 according the framework in Appendix D, to find relevant features to use for the classifier. Based on this, we trained a classifier using the following nodes/features: PrimaryTumor, CA125, Therapy, Recurrence. The input file for the  ODD compiler is shown on Figure 10.\\
    
    We now consider how 5-year survival can be improved using interventions. In Figure 13, we use the IntRob software to assess the maximal probability of 5-year survival, under all possible interventions to recurrence and adjuvant therapy. We find in Figure 13 that the upper bound on this probability is 1 (up to floating point error), indicating that there may be significant room for improvement of patient outcomes by improving treatment policies and reducing recurrence. 
    %The process of bounds finding for third experiment is shown in Fig.12 and Fig.13.
    \end{figure}
    \begin{figure}[htb]
       \centering
        \includegraphics[width=1\linewidth, angle=0]{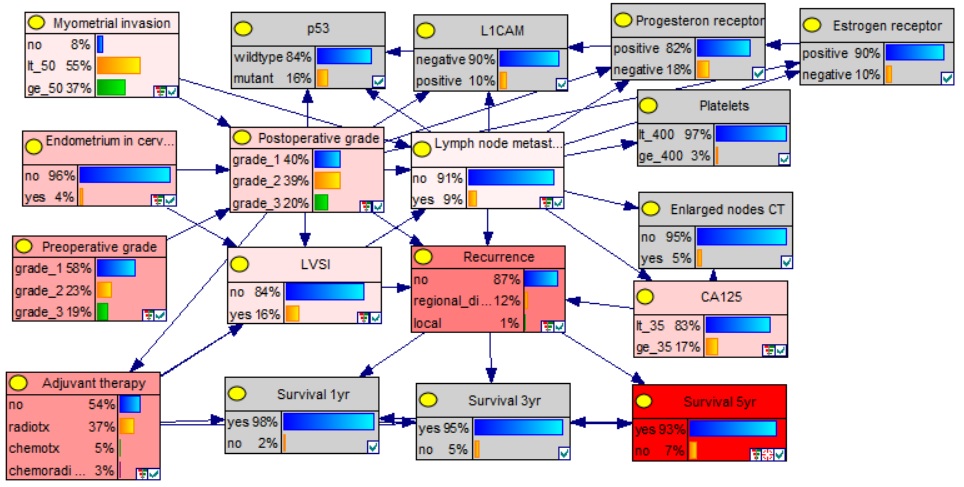}
         \caption{Fig. 9 Sensitivity analysis}
          \label{fig:e9}
    \end{figure}
    \begin{figure}
    %The ordered decision diagram was made for the followings nodes:
    %PrimaryTumor, CA125, Therapy, Recurrence. The input file for ODD compiler is shown on Fig.10
    \end{figure}
    
    \begin{figure}[htb]
       \centering
        \includegraphics[width=0.65\linewidth, angle=0]{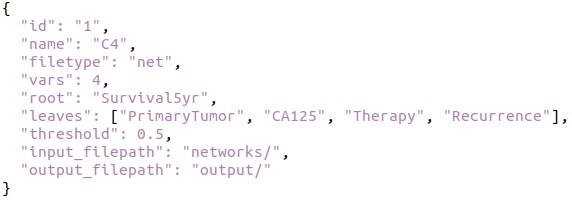}
         \caption{Fig. 10 JSON config file for 5-year survival classifier}
          \label{fig:e10}
    \end{figure}
    \begin{figure}
    The output file after compiling in BNC\_SDD software is shown in Figure 11.
    \end{figure}
    
    \begin{figure}[htb]
       \centering
        \includegraphics[width=0.5\linewidth, angle=0]{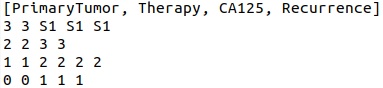}
         \caption{Fig. 11 ODD file for 5-year survival classifier}
          \label{fig:e11}
    \end{figure}
    \begin{figure}
    
    \end{figure}
    \begin{figure}[htb]
       \centering
        \includegraphics[width=1\linewidth, angle=0]{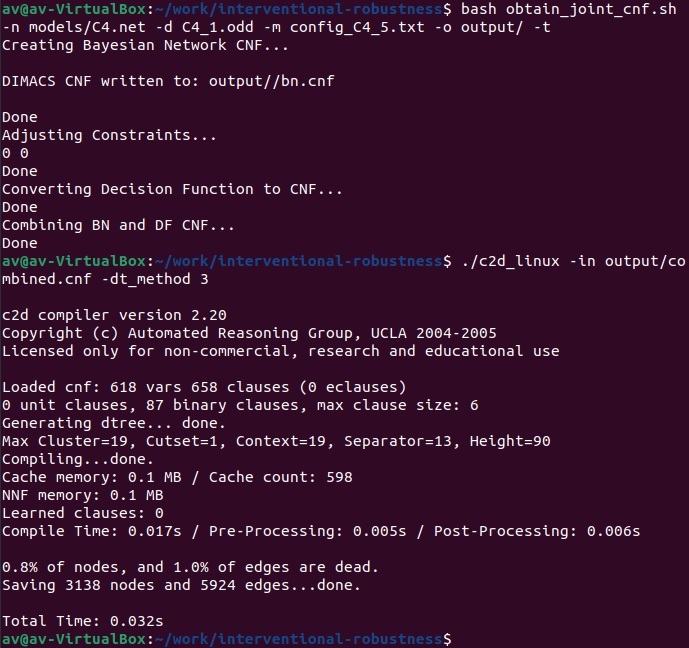}
         \caption{Fig. 12 IntRob software for the final experiment}
          \label{fig:e12}
    \end{figure}
    \begin{figure}[htb]
       \centering
        \includegraphics[width=1\linewidth, angle=0]{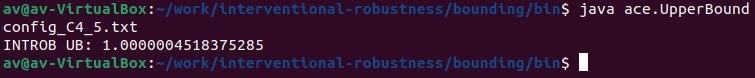}
         \caption{Fig. 13 Upper bound on 5-year survival probability under best-case intervention}
          \label{fig:e13}
    \end{figure}
    
    \newpage
    \clearpage
    %The used model for experiments was created for the tasks of counselling patients and joint decision-making \cite{reijnen_preoperative_2020}.
    %As a result of the software operation, an upper limit of interventions was obtained. In the first experiment the probability of falling into the risk group during therapy is 59.1\%. 
    %For second one the upper bound of the probability of getting a recurrence was 37.7\%. 
    
    %In the third experiment we received a probability more than one. It is related to a IEEE 754 technical standard for floating-point computation and has nothing to do with the operation of the software.Thus, we can say that the upper bound of probability is 100\%.
    
    In conclusion, our analysis has revealed a number of insights into the Bayesian network model. The obtained BN bounds can be utilised to develop subsequent treatments strategies, as well as used as input into models that improve QoL or QoL \cite{xu_modeling_2018} or facilitate EOL at home \cite{kern_impact_nodate}. It is also possible to use medical protocols that reduce the likelihood of falling into a certain group.

\subsubsection*{Acknowledgements}

This project was funded by the ERC under the European Union’s Horizon 2020 research and innovation programme (FUN2MODEL, grant agreement No.834115).

\newpage
\bibliographystyle{ieeetr}
\bibliography{report_edited}

\addcontentsline{toc}{section}{References}

\newpage
\addcontentsline{toc}{section}{Appendix A}

\includepdf[pages={1-3},fitpaper,rotateoversize, pagecommand=\thispagestyle{plain}]{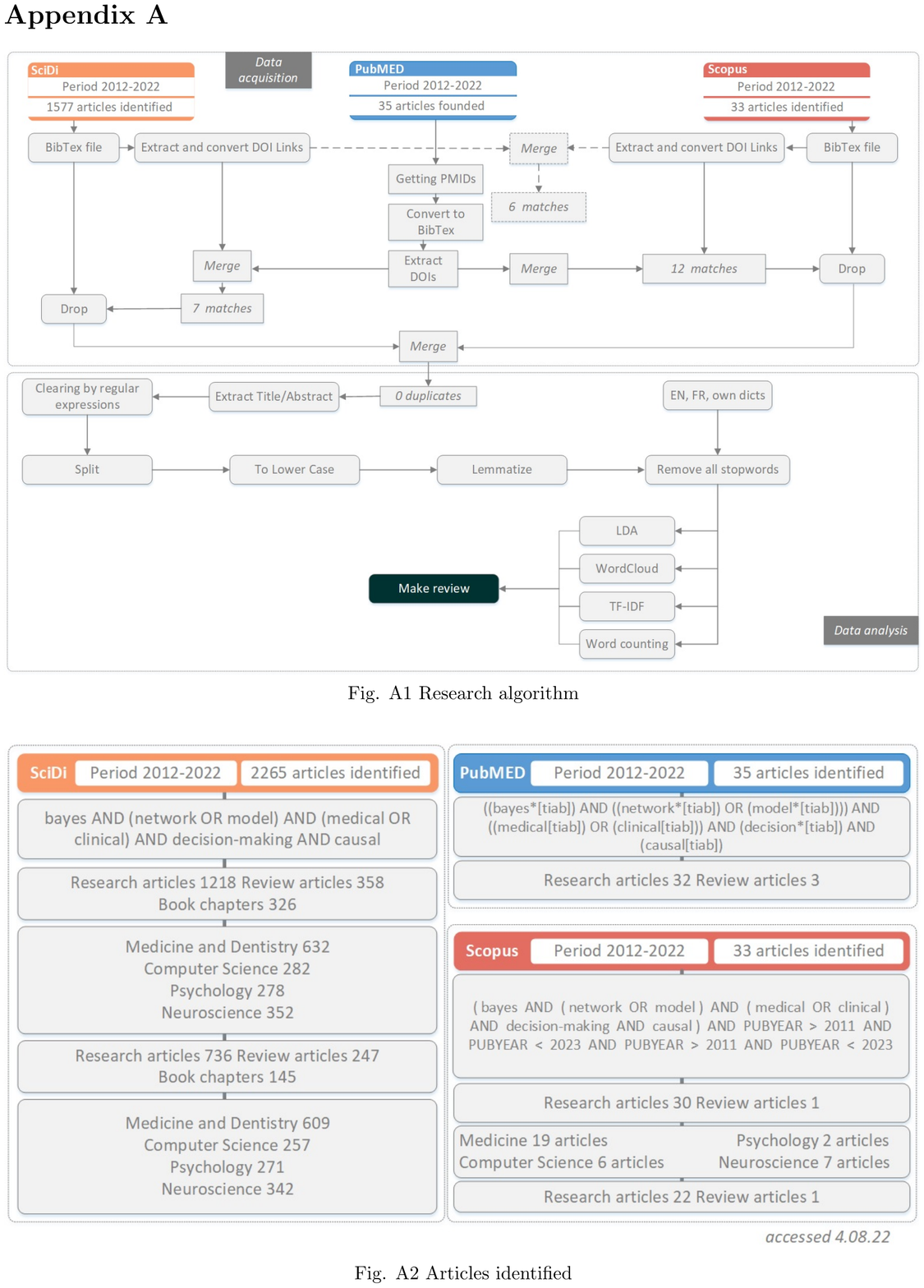}
\includepdf[pages={4-12}, fitpaper,rotateoversize,turn=true]{pdf/app_a.pdf}

\addcontentsline{toc}{section}{Appendix B}
\includepdf[pages={-},fitpaper,rotateoversize, pagecommand=\thispagestyle{plain}]{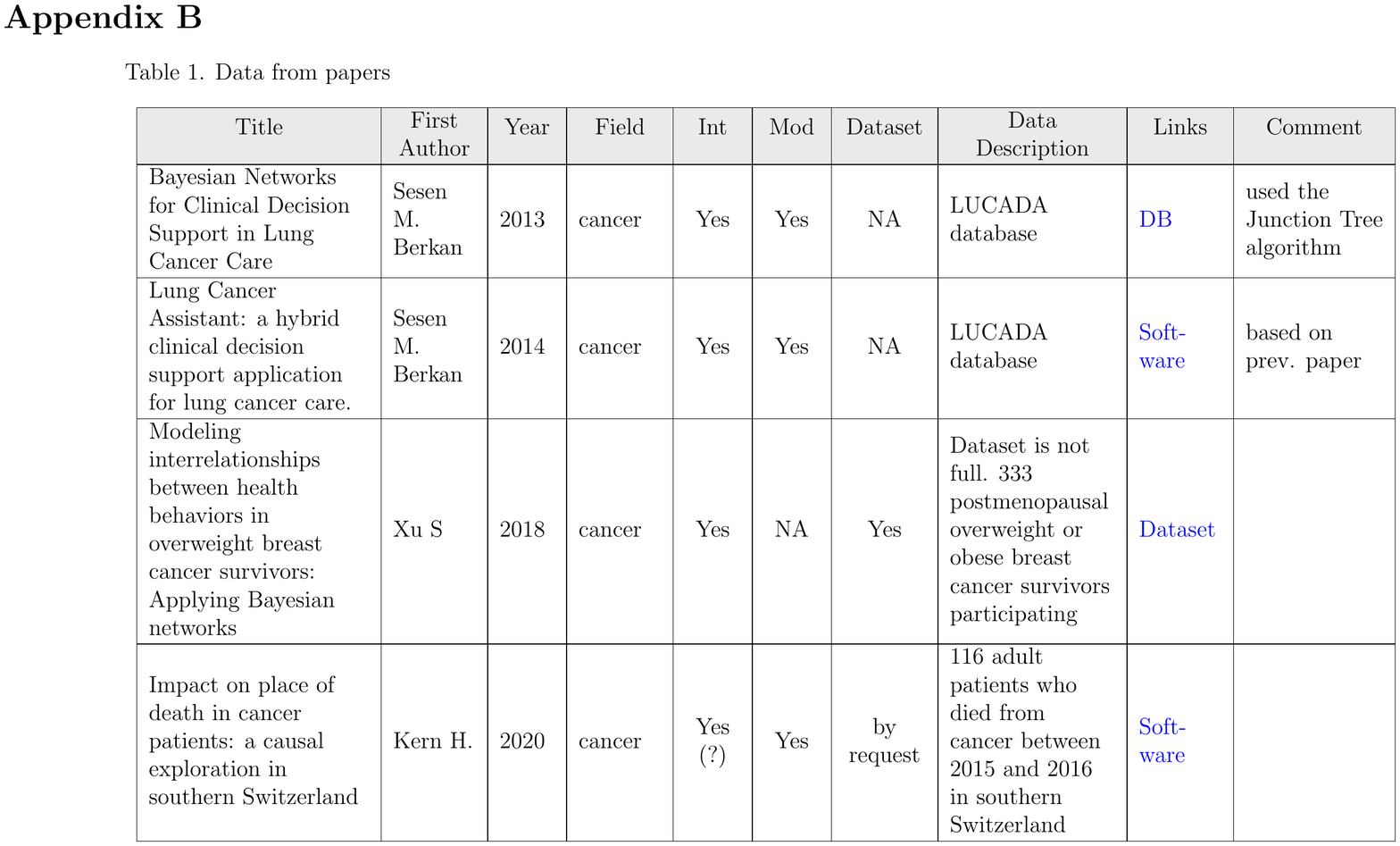}
\addcontentsline{toc}{section}{Appendix C}
\includepdf[pages={-},fitpaper,rotateoversize, pagecommand=\thispagestyle{plain}]{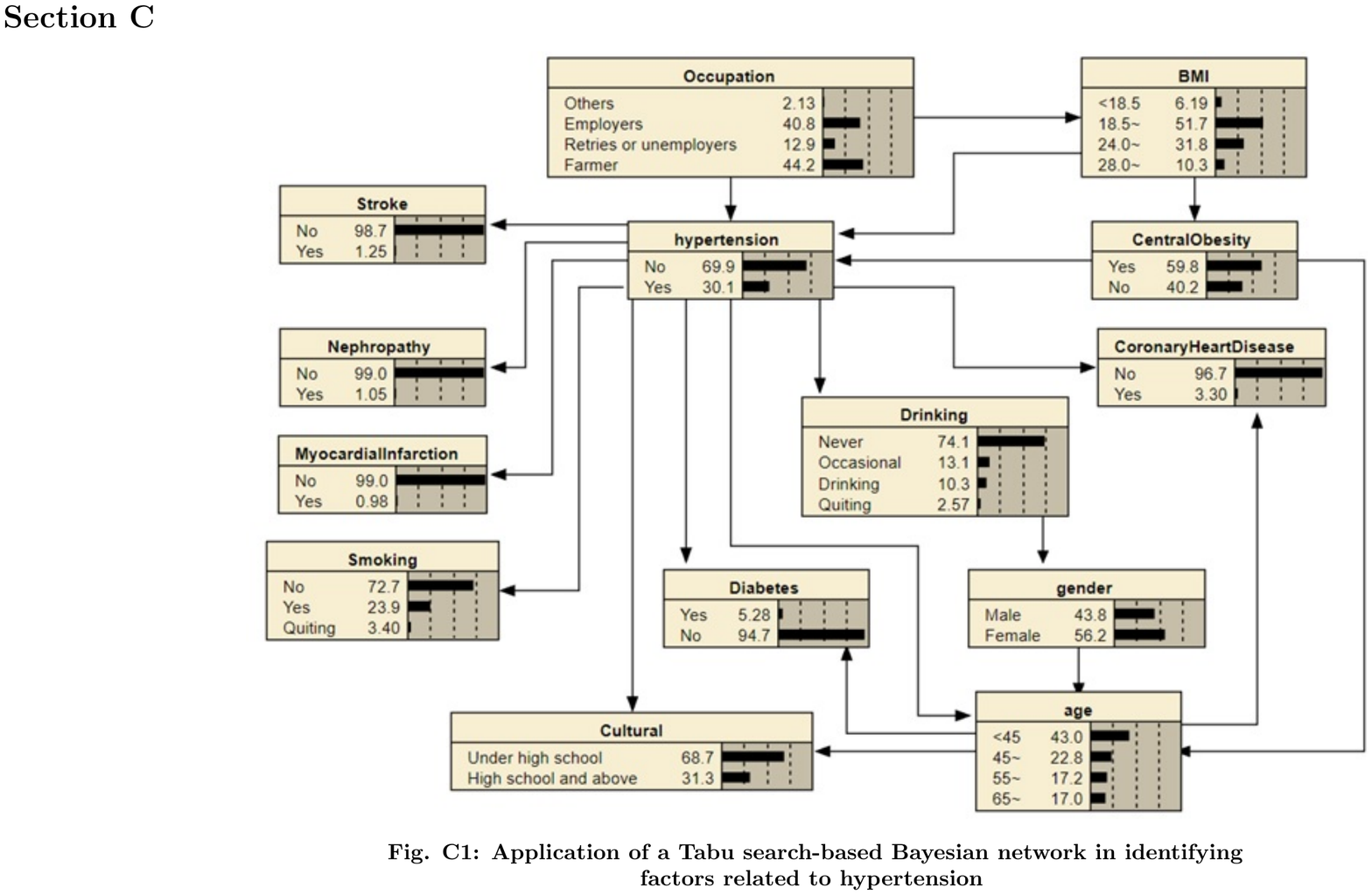}
\addcontentsline{toc}{section}{Appendix D}
\includepdf[pages=-,pagecommand=\thispagestyle{plain}]{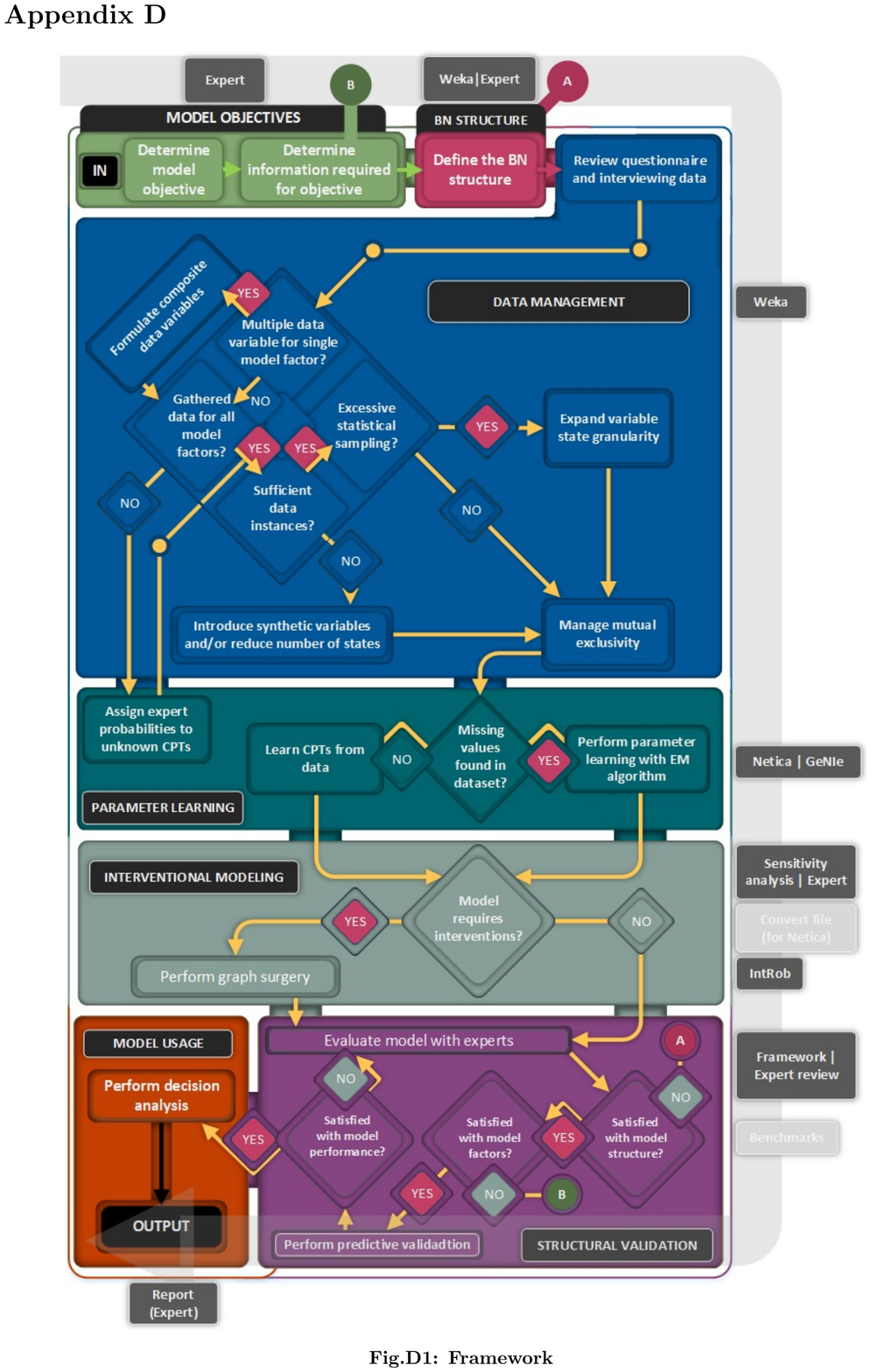}
%\addcontentsline{toc}{section}{Appendix E}
%\includepdf[pages=-,pagecommand=\thispagestyle{plain}]{pdf/app_e.pdf}

\end{document}